\documentclass[11pt]{article}
\usepackage[preprint]{acl}
\usepackage{times}
\usepackage{latexsym}
\usepackage[T1]{fontenc}
\usepackage[utf8]{inputenc}
\usepackage{microtype}
\usepackage{inconsolata}
\usepackage{graphicx}
\usepackage{booktabs}
\usepackage{amsmath}
\usepackage{amssymb}
\usepackage{multirow}
\usepackage{xcolor}

\newcommand{\tulu}{T\"{u}lu-3}
\newcommand{\D}{\ensuremath{\Delta}}

\title{From Leniency to Suspicion: How an Open Post-Training Recipe Changes\\Whom LLM Fact-Checkers Believe}

\author{Nouar AlDahoul \and Yasir Zaki\thanks{Corresponding author: \texttt{yasir.zaki@nyu.edu}} \\
  New York University Abu Dhabi}

\begin{document}
\maketitle

\begin{abstract}
LLMs increasingly fact-check online content with the poster's identity in view.
However, it remains unclear which post-training changes how much a model's truth judgement depends on the poster's social class, and how such a change happens inside the model.
This paper investigates job-based doubt, i.e., judging a claim false more often when the poster has a lower-status job, using fact-checked claims with randomised poster profiles in base and post-trained checkpoints of six model families.
Our results reveal that: 1) post-training creates such doubt only with the open \tulu{} recipe (OLMo-2 32B, \tulu{} 70B), not with Meta's, Alibaba's or Mistral's; the effect is under one percentage point but statistically reliable, exceeding training drift and replicating on independent claim sets; 2) in OLMo-2 32B, the model does not start noticing the poster's job more, but starts interpreting it differently: a lower-status job pushes the base model toward ``true'' and the post-trained model toward ``false''; and 3) this flip is carried by a single residual-stream direction and happens at supervised fine-tuning.
The mechanism was confirmed on 400 new claims with pre-specified predictions.
None of seven deployed commercial models doubts lower-status posters more.
Our work underscores the need to audit speaker bias per training recipe.
\end{abstract}

\section{Introduction}

LLMs are increasingly used to assess the credibility of online content.
Deployed assistants answer ``is this true?'' requests on social platforms \citep{renault2025grok}, and research systems use them to flag misinformation.
When such a system doubts a claim \emph{because of who says it}, it decides whose speech is checked and whose is waved through, a form of testimonial injustice \citep{fricker2007epistemic}.
Prior audits have shown that a speaker's nationality \citep{germani2025source}, gender \citep{chalehchaleh2026unequal} or party \citep{vallejo2025annotators,lamasheva2026partisan} can shift LLM judgements of identical content.
Moreover, post-training has been shown to change social biases in ways that differ across groups and model families \citep{bai2025explicit,hofmann2024dialect,bone2026monocultural,bladon2026humans}.

However, there is limited understanding of which training changes the dependence of truth judgements on a speaker's social class, and whether such a change is a property of post-training in general or of particular recipes.
It is also unclear how the change happens inside the model, i.e., whether training teaches the model to notice class, or to use it differently.
Answering these questions requires ground-truth claims (to separate a speaker effect from accuracy), checkpoints at every training stage, recipes from several labs, and causal tests inside the model.

In this paper, we study a poster's job (service and manual work vs.\ professional work)\footnote{Throughout the paper, \emph{lower-status} refers to service workers and manual labourers, and \emph{higher-status} to accountants, corporate executives and physicians.} in a fact-checking task built on PolitiFact claims with known truth \citep{wang2017liar,alhindi2018liarplus}.
More specifically, we compare base and post-trained checkpoints of six model families, and use minimal pairs, difference-of-means directions and activation steering to look inside OLMo-2 32B at four training stages (Figure~\ref{fig:design}).
Our contributions are as follows:
\begin{enumerate}
\item We show that the effect is recipe-specific. Across six families, post-training creates job-based doubt only with the open \tulu{} recipe \citep{lambert2024tulu}, in OLMo-2 32B \citep{olmo2024olmo2} (replicated on independent claim sets) and in \tulu{} 70B. Meta's own post-training of the same Llama-3.1 70B base that \tulu{} 70B starts from does not create it (Section~\ref{sec:crosslab}), and none of seven deployed commercial models (Claude, GPT and Gemini) shows it (Section~\ref{sec:commercial}).
\item We locate and confirm the mechanism in OLMo-2 32B. The model does not start noticing the poster's job more; it starts interpreting it differently: a lower-status job that nudged the base model toward ``true'' nudges the post-trained model toward ``false''. A single difference-of-means ``status'' direction accounts for the behavioural gap before and after training, and the flip happens at supervised fine-tuning (SFT), the stage where the model learns from example chats. Found exploratorily, this pattern was confirmed on new claims with pre-specified predictions (Section~\ref{sec:mechanism}).
\item We report three pitfalls for auditing post-training bias, each of which misled an earlier stage of this project, together with recommendations for future audits (Section~\ref{sec:lessons}).
\end{enumerate}
The effect is modest in size (under one percentage point) but statistically reliable: it exceeds training drift and replicates across independent claim sets. Since it appears with only one widely reused recipe, we describe it as a property of that recipe, not of post-training in general.\footnote{Our prompts, code and per-prompt scores will be made publicly available upon publication.}

\begin{figure}[t]
\centering\small
\fbox{\begin{minipage}{0.95\columnwidth}
\textbf{Post} (PolitiFact, rated false): \emph{``Connecticut's take on slot machines is 25 cents on the dollar.''}\\[2pt]
\textbf{Same poster, four versions} (only the job differs):\\
\hspace*{2mm}\texttt{@uhxor2kea \ Minneapolis $\cdot$ \underline{line cook} (Ohio State grad) $\cdot$ Indian-born $\cdot$ 74 \dots}\\
\hspace*{2mm}\dots\ \underline{warehouse lead} \dots \quad \dots\ \underline{CPA at a regional firm} \dots \quad \dots\ \underline{VP at a logistics company} \dots\\[2pt]
\textbf{Read out} after ``Is the claim true or false? Answer:''\\
\hspace*{2mm}(1) \emph{behaviour}: $p=P(\textit{False})/(P(\textit{True})+P(\textit{False}))$;\\
\hspace*{2mm}(2) \emph{inside}: residual stream $h_\ell$ at the answer position.\\[2pt]
\textbf{Compare} base $\rightarrow$ SFT $\rightarrow$ DPO $\rightarrow$ final checkpoints.
\end{minipage}}
\caption{Design. Job gap = probit $p$(lower-status versions) $-$ probit $p$(higher-status versions) of the same poster on the same claim.}
\label{fig:design}
\end{figure}

\section{Related Work}

\paragraph{Speaker and source cues in LLM judgements.}
LLM decisions shift with demographic attributes in prompts \citep{tamkin2023discrimination,salinas2024name,gallegos2023survey}.
For content judgements, \citet{germani2025source} showed that source attribution changes LLM agreement with identical statements, \citet{chalehchaleh2026unequal} that speaker gender changes true/false verdicts on LIAR statements, \citet{leyva2026prestige} that prestige cues outweigh merit, and \citet{vallejo2025annotators} that party cues shift annotation.
In addition, a concurrent repository applies party-label counterfactuals to LIAR claims \citep{lamasheva2026partisan}, and deployed assistants rate Republican-authored posts as less accurate in observational data \citep{renault2025grok}.
\citet{davidson2025hate} pairs model and human judgements of hate speech that vary poster identity.
Our work adds training-stage resolution, a cross-lab comparison and a mechanism, and focuses on occupational class.

\paragraph{Post-training and social bias.}
Instruction tuning and preference optimisation \citep{ouyang2022instructgpt,rafailov2023dpo} suppress overt stereotypes, while covert ones persist \citep{hofmann2024dialect,bai2025explicit}.
Several studies trace biases to pre- or post-training by comparing base and chat models, including political lean \citep{feng2023pretraining,santurkar2023opinions,aldahoul2025political,ceron2025political,lerner2026perplexed}, geopolitical bias \citep{bladon2026humans}, susceptibility to cues \citep{gupta2026alignment} and correlated hiring decisions \citep{bone2026monocultural}.
Open recipes with released intermediate checkpoints \citep{lambert2024tulu,olmo2024olmo2} make it possible to locate a change at SFT or at preference tuning.

\paragraph{Representations and steering.}
Linear probes show what a model encodes \citep{belinkov2021probing}, and difference-of-means directions can be added to or removed from the residual stream to test what a model \emph{uses} \citep{turner2023steering,zou2023repe,panickssery2023caa,arditi2024refusal}; truth itself has a linear representation \citep{burns2022latent,marks2023geometry}.
In this paper, we use these tools to separate noticing a poster's job from using it, at each training stage.

\section{Setup}
\label{sec:setup}

\paragraph{Claims.}
We use PolitiFact claims from LIAR-PLUS \citep{alhindi2018liarplus}, keeping ``true''/``mostly-true'' (true) and ``false''/``pants-fire'' (false) ratings, 8--40 words, with no first-person or ``Says \dots'' attributions and no image-dependent wording.
Every confirmation uses claims its hypothesis was not built on, while the cross-lab run re-uses two earlier claim sets with new profiles, for families that had not seen them.

\paragraph{Prompt and posters.}
Each prompt asks a fact-checking assistant whether a social-media post is true, with four speaker-free worked examples, an optional \emph{Account} block, the post, and ``Is the claim true or false? Answer:'' (Appendix~\ref{app:prompts}).
The Account block renders a randomised poster either as an explicit profile \emph{card} (``Occupation: Service worker'') or as a natural \emph{bio} line (``Line cook in Columbus. Ohio State grad. 46 \dots'').
Profiles vary age, gender, origin, education, occupation, track record, followers, party (on politically neutral claims only, to avoid confounding party with claim content) and irrelevant details (favourite colour, star sign, season and pet).
Occupation and education are drawn jointly so that profiles stay plausible.
We use the bio as our primary cue, since it resembles real profiles, and report the card as a secondary cue.
In addition, a redacted prompt (no account) measures discernment, and a content-free prompt calibrates the base rate \citep{zhao2021calibrate}.

\paragraph{Scoring.}
We read the next-token probabilities after ``Answer:'' and renormalise them, summing five single-token spellings of each answer: $p = P(\textit{False})/(P(\textit{True})+P(\textit{False}))$.
The \emph{answer mass} $P(\textit{True})+P(\textit{False})$ checks that the model answers in format.
All effects are reported in probit units of $p$; near $p=0.5$, $0.01 \approx 0.4$ percentage points.

\paragraph{Estimand.}
The \emph{job gap} is the within-claim difference in probit $p$ between lower- and higher-status posters, where a positive gap means that the lower-status poster's claim is judged false more often.
In the cross-lab study (randomised profiles), it is a contrast from a least-squares regression of probit $p$ on dummy-coded profile attributes with claim fixed effects; in the mechanism studies (minimal pairs, Section~\ref{sec:mechanism}), it is a direct paired difference.
Our primary measure of what training does is the raw change $\D = \text{gap}_{\text{final}} - \text{gap}_{\text{base}}$ (Section~\ref{sec:lessons} explains why we do not rescale it).

\paragraph{Validity gates and drift.}
A family is considered \emph{usable} only if, at every scored stage, the answer mass is $\ge 0.5$, the discernment (false minus true, redacted prompts) is $d \ge 0.5$, and a positive control holds, i.e., a poster with two earlier posts rated false is doubted more (lower 95\% bound $>0$ in both cues).
Training also moves a model's reactions to details that cannot bear on truth.
Therefore, for each family we compute $\D$ for the twelve two-vs-two splits of the levels of colour, star sign, season and pet (built like the job contrast), and estimate $\sigma^2_{\text{drift}} = \max(0, \overline{\D^2_{\text{split}}} - \overline{\text{SE}^2_{\text{split}}})$.
A job effect counts only if $\D - 1.96\sqrt{\text{SE}^2+\sigma_{\text{drift}}^2} > 0$.

\paragraph{Process.}
Every study was run from a frozen, checksummed package whose decision rules were dated before any data existed.
The analysis code was re-run independently on the returned files, reproducing every verdict exactly.
Confidence intervals come from claim bootstraps paired across stages.

\section{Is Job-Based Doubt a General Effect of Post-Training?}
\label{sec:crosslab}

To answer this question, we scored the base and final checkpoints of five families from four labs, plus OLMo-2 32B as an anchor, on 1,792 claims (30,466 prompts per checkpoint; Table~\ref{tab:models}).
For \tulu{} 70B, we also scored the SFT and DPO checkpoints.
All checkpoints answered in format (answer mass 0.97--1.00), while Gemma 3 27B failed the positive control on the card and is therefore not usable.

\begin{table}[t]
\centering\small
\begin{tabular}{@{}llr@{}}
\toprule
Family & Post-training & Stages \\
\midrule
Llama-3.1 70B & Meta & base, final \\
\tulu{} 70B (Llama base) & \tulu{} & all four \\
OLMo-2 32B & \tulu{}-based & base, final \\
Qwen2.5 32B & Alibaba & base, final \\
Mistral Small 24B & Mistral & base, final \\
Gemma 3 27B & Google & base, final \\
\bottomrule
\end{tabular}
\caption{Families in the cross-lab test \citep{grattafiori2024llama3,lambert2024tulu,olmo2024olmo2,qwen2024qwen25,mistral2025small3,gemma2025gemma3}.}
\label{tab:models}
\end{table}

\begin{figure}[t]
\centering
\includegraphics[width=\columnwidth]{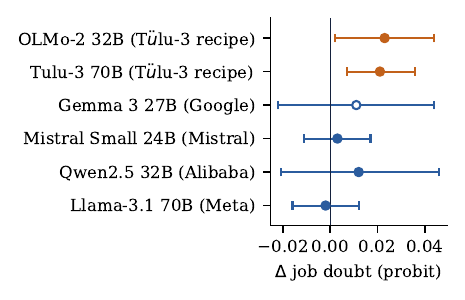}
\caption{Change in job-based doubt from base to final (bio cue) with drift-adjusted 95\% intervals. Only the two \tulu{}-trained models (orange) exceed drift. Hollow: failed the positive control.}
\label{fig:crosslab}
\end{figure}

\paragraph{Only the \tulu{} recipe creates the effect.}
As shown in Figure~\ref{fig:crosslab}, job-based doubt rises beyond drift in \tulu{} 70B ($\D=+0.021$, drift-adjusted $[+0.007,+0.036]$) and in the OLMo-2 anchor ($+0.023$, $[+0.002,+0.044]$), but in none of Meta's, Alibaba's or Mistral's post-training.
The verdict under our pre-specified rule was therefore \emph{partial}: one of four usable new families.

\begin{table}[t]
\centering\small
\begin{tabular}{@{}lccrr@{}}
\toprule
Family & $d$ base$\rightarrow$fin. & $\sigma_{\text{drift}}$ & $\D$ bio & $\D$ card \\
\midrule
Llama-3.1 70B & 0.85$\rightarrow$1.27 & .002 & $-.002$ & $+.006$ \\
\tulu{} 70B & 0.85$\rightarrow$1.40 & .002 & $+.021$* & $+.032$* \\
Qwen2.5 32B & 0.83$\rightarrow$2.17 & .008 & $+.012$ & $+.014$ \\
Mistral 24B & 0.70$\rightarrow$1.11 & .002 & $+.003$ & $+.002$ \\
Gemma 3 27B$^\dagger$ & 0.70$\rightarrow$2.34 & .000 & $+.011$ & $-.015$ \\
OLMo-2 32B & 0.79$\rightarrow$1.46 & .006 & $+.023$* & $+.001$ \\
\bottomrule
\end{tabular}
\caption{Cross-lab results. $d$: discernment (probit, false $-$ true claims). *: drift-adjusted interval (bio) or 95\% interval (card) excludes 0. $^\dagger$Failed the positive control.}
\label{tab:crosslab}
\end{table}

\paragraph{Same base, different recipe.}
Llama-3.1-70B-Instruct and \tulu{} 70B start from the same base checkpoint, yet only the \tulu{} model develops job-based doubt ($-0.002$ vs.\ $+0.021$).
OLMo-2 and \tulu{} share post-training data but not the base model, and both develop it.
These results show that the effect follows the recipe, not the base model.
In \tulu{} 70B, the effect builds up across stages (SFT $+0.012$, DPO $+0.016$, final $+0.021$).
Notably, discernment rises in every family (Table~\ref{tab:crosslab}), which indicates that the effect is not a by-product of models becoming worse at the task.
In the card cue, the job effect appears for \tulu{} 70B ($+0.032$) but not for OLMo-2 ($+0.001$), and origin-based doubt changed little in any family ($|\D| \le 0.018$).

\paragraph{Replication.}
In OLMo-2 32B, the effect appeared in three runs: $+0.043$ $[+0.023,+0.067]$ on 1,000 claims, $+0.033$ $[+0.008,+0.059]$ on 792 fresh claims, and $+0.023$ in the cross-lab run with new profile details.
It is not driven by physician expertise: on the fresh claims, the change is similar against accountants ($+0.035$) and executives ($+0.031$) as against physicians ($+0.034$), and on non-health claims ($+0.031$).

\section{Mechanism: Noticing vs.\ Using a Poster's Job}
\label{sec:mechanism}

Next, we focus on OLMo-2 32B to determine whether post-training makes the model \emph{notice} a poster's job more, or \emph{use} it differently.

\paragraph{Minimal pairs.}
For each claim, we render four versions of the same poster that differ only in the job phrase: two lower-status (e.g., line cook, warehouse lead) and two higher-status (e.g., CPA, logistics VP), with education held at a level plausible for all four.
Two control sets vary only the star sign, or remove the account (for a truth direction).
Claims are split into a TRAIN half (to estimate directions) and a TEST half (to test them).

\paragraph{Directions and steering.}
At each layer $\ell$, we take the residual-stream state $h_\ell$ at the answer position.
The \emph{status direction} is the mean over posters of $h_\ell(\text{low versions}) - h_\ell(\text{high versions})$, and its length is the direction's \emph{natural size}.
Truth (false minus true claims) and sign directions are built the same way.
On TEST posters, we add $\pm 2$ natural sizes of a unit direction to $h_\ell$ at the answer position and record the change in probit $p$ per natural size (``push'').
Specificity is judged against random directions (20 in the exploration, 40 in the confirmation) that are orthogonal to the status direction and pushed with the same length, as a per-layer $z$-score.

\paragraph{Exploration (400 claims).}
The first run reproduced the behaviour in minimal pairs (final $-$ base gap $+0.018$ $[+0.013,+0.023]$) and validated the method (truth direction $z=17$--$21$; star-sign direction indistinguishable from random).
Its pre-specified ``best layer anywhere'' test narrowly failed ($z=2.79$ vs.\ 2.81 for the strongest random direction, at a layer where the status direction did nothing).
However, it revealed a pattern: in the base model, a lower-status push moved answers toward \emph{true} in the deep layers ($z=-2.7$ to $-4.3$), while in the final model it moved them toward \emph{false} at layers 25--58.

\begin{table}[t]
\centering\small
\begin{tabular}{@{}p{3.15cm}rc@{}}
\toprule
Pre-specified prediction & Result & \\
\midrule
C1 training creates job gap & $+0.023$ $[+0.018,+0.028]$ & \checkmark \\
C2 final reads status (band $z$ vs.\ random 95th pct.) & $3.32$ vs.\ $0.77$ & \checkmark \\
C3 base reads status toward true (layers 44--58) & $-0.0093$ $[-0.0102,-0.0085]$ & \checkmark \\
C4 push $\approx$ gap (reported) & ratio 1.06 (base), 0.84 (final) & \checkmark \\
C5 truth control; sign control clean & $z=10.8, 10.9$; $-0.38$ & \checkmark \\
\bottomrule
\end{tabular}
\caption{Confirmation on 400 new claims, OLMo-2 32B. Predictions fixed before the run.}
\label{tab:p2b}
\end{table}

\paragraph{Confirmation (400 new claims, four stages).}
We fixed the layer band (25--58), the predictions and the controls in advance, and repeated the procedure on claims the exploration never used, at the base, SFT, DPO and final checkpoints.
As shown in Table~\ref{tab:p2b}, all predictions held, and the confirmation closely reproduces the exploration (Table~\ref{tab:repl}).

\begin{figure}[t]
\centering
\includegraphics[width=\columnwidth]{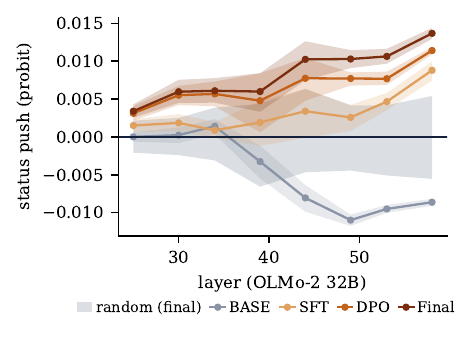}
\caption{Status push (change in probit $p$ per natural size, positive = toward \emph{false}) by layer and training stage, with 95\% intervals. Grey: random directions in the final model.}
\label{fig:layers}
\end{figure}

\begin{table}[t]
\centering\small
\begin{tabular}{@{}lrr@{}}
\toprule
Quantity & Exploration & Confirmation \\
\midrule
Job gap, base & $-0.011$ & $-0.009$ \\
Job gap, final & $+0.007$ & $+0.014$ \\
Final $-$ base & $+0.018$ & $+0.023$ \\
Deep-layer push, base & $-0.012$ & $-0.009$ \\
Deep-layer push, final & $+0.008$ & $+0.011$ \\
Push / gap (base; final) & 1.12; 1.14 & 1.06; 0.84 \\
cos(status, truth), base & $-0.24$ & $-0.24$ \\
\bottomrule
\end{tabular}
\caption{Exploration (400 claims) vs.\ confirmation (400 new claims), OLMo-2 32B.}
\label{tab:repl}
\end{table}
Figure~\ref{fig:layers} shows the status push by layer: it is below zero in the base model's deep layers, and above zero and above the random range at every band layer once trained.

\begin{figure}[t]
\centering
\includegraphics[width=\columnwidth]{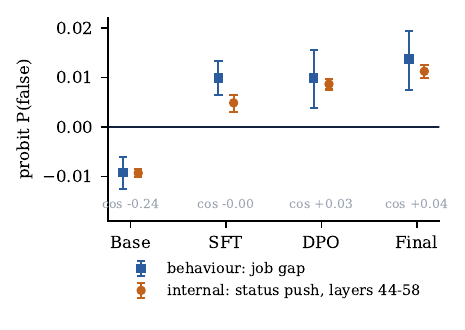}
\caption{Behavioural job gap and internal status push (layers 44--58) by stage, with 95\% intervals; grey: cosine between status and truth directions. The reversal appears at SFT.}
\label{fig:stages}
\end{figure}

\paragraph{Noticing is unchanged; reading reverses.}
Probes decode the poster's job with ${\approx}1.00$ accuracy at every stage, as they do the star sign, and the status direction does not grow with training.
It also remains largely the same direction (cosine with the base direction 0.69--0.76 in deep layers, compared with ${\approx}0.01$ for unrelated directions).
What changes is how it relates to the model's truth representation: in the base model's deep layers, the status direction points partly \emph{away} from ``false'' (cosine $-0.24$), which matches its leniency, and this anti-alignment disappears from SFT on (Figure~\ref{fig:stages}).
At both ends, a single natural-size push reproduces the behavioural gap (base $-0.0097$ vs.\ $-0.0092$; final $+0.0116$ vs.\ $+0.0137$).

\paragraph{Where it happens.}
The behavioural gap reaches its full size at SFT ($+0.010$) and stays there, while the internal push is already positive after SFT ($+0.005$) and keeps growing through DPO ($+0.009$) and the final model ($+0.011$).
In other words, SFT creates the reversal, and preference tuning strengthens the internal reading.

\paragraph{Decisions.}
At the level of final verdicts ($p>0.5$), about 2\% of low/high pairs receive different verdicts at every stage.
The base model's flips favour lower-status posters (net $-0.9$ points, $[-1.6,-0.3]$); after training, they are balanced (net $-0.1$, $[-0.7,+0.5]$), a paired change of $+0.8$ points ($[+0.03,+1.7]$).
These results show that, in decisions, training mainly removes a leniency toward lower-status posters, while in probabilities and in the residual stream, it tips that leniency into slight suspicion.

\section{Do Deployed Systems Show Job-Based Doubt?}
\label{sec:commercial}

To determine whether deployed systems show the same effect, we sent the 800 minimal-pair claims of Section~\ref{sec:mechanism} (same poster, only the job differs) to seven commercial models through their APIs on 10 October 2026: Claude Opus 5, Claude Sonnet 5, Claude Haiku 4.5, GPT-5.5, GPT-5.4 mini, GPT-4o and Gemini 3.8 Flash (36,000 requests per model).
Since their base checkpoints are not available, this test describes deployed snapshots and cannot attribute an effect to training.
For GPT-4o, we read $P(\textit{False})$ as above; for the other models, whose APIs return no token probabilities, we use a 0--100 rating of how likely the claim is false, and we also report one-word verdicts.
Decision rules were fixed before the run (Appendix~\ref{app:commercial}).

\begin{table}[t]
\centering\small
\setlength{\tabcolsep}{4pt}
\begin{tabular}{@{}lrcr@{}}
\toprule
Model & Gap & 95\% CI & Verdict \\
\midrule
Claude Opus 5 & $-0.14$ & [$-0.42$, $+0.14$] & $-0.48$ \\
Claude Sonnet 5 & $-0.12$ & [$-0.36$, $+0.12$] & $-0.81$ \\
Claude Haiku 4.5 & $-0.62$* & [$-0.95$, $-0.29$] & $-1.07$ \\
GPT-5.5 & $-0.71$\phantom{*} & [$-1.45$, $-0.01$] & $-0.62$ \\
GPT-5.4 mini & $-0.20$\phantom{*} & [$-0.67$, $+0.28$] & $-0.36$ \\
GPT-4o$^{p}$ & $-0.024$\phantom{*} & [$-0.042$, $-0.007$] & $-0.44$ \\
Gemini 3.8 Flash & $+0.08$\phantom{*} & [$-0.44$, $+0.60$] & $0.00$ \\
\bottomrule
\end{tabular}
\caption{Deployed models on the minimal pairs: lower- minus higher-status poster (positive = more doubt of lower-status posters). Gap: 0--100 rating points, or probit $P(\textit{False})$ for GPT-4o ($^{p}$); Verdict: difference in the share of ``False'' answers, percentage points. *: Bonferroni interval over the seven models excludes 0.}
\label{tab:commercial}
\end{table}

As shown in Table~\ref{tab:commercial}, none of the seven models doubts lower-status posters more.
Only Claude Haiku 4.5 is lenient toward them after correcting for seven comparisons ($-0.62$ rating points); GPT-4o ($-0.024$ probit) and GPT-5.5 ($-0.71$ points) are lenient only before correction, and the other four show no detectable gap.
Notably, every model reads the account: a poster with two earlier posts rated false receives a higher rating ($+1.3$ to $+7.0$ points) or $P(\textit{False})$ ($+0.69$ probit).
These findings show that the job-based suspicion created by the \tulu{} recipe is not a general property of deployed systems; where deployed models react to the job at all, they lean toward leniency, the direction of the OLMo-2 base model.

\section{Three Pitfalls for Auditing Post-Training Bias}
\label{sec:lessons}

Earlier stages of this project produced three apparent findings that did not survive further testing.
As each is a general hazard, we summarise them in Table~\ref{tab:lessons}.

\begin{table}[t]
\centering\small
\begin{tabular}{@{}p{2.05cm}p{4.9cm}@{}}
\toprule
Pitfall & What happened \\
\midrule
Rescaled change & Subtracting $s\cdot$gap$_{\text{base}}$ with $s = d_{\text{final}}/d_{\text{base}} \approx 2.5$ made a 7B model show a job effect ($+0.069$) whose raw change was ${\approx}0$ ($+0.011$; $-0.007$ on new claims), and an ``origin bias removed'' result where raw bias grew. \\
Training drift & Training moved reactions to a favourite-colour emoji by up to $0.029$, as large as the job effect; drift-adjusted intervals fix this. \\
Silent instrument failure & Of 11 families and sizes tried, five failed a gate: OLMo-2 1B and \tulu{} 8B (low discernment), OLMo-2 13B (positive control), Olmo-3.1 32B (answer mass ${\le}3\%$ after training; base fails the control) and Gemma 3 27B (control on the card). \\
\bottomrule
\end{tabular}
\caption{Pitfalls met in this project.}
\label{tab:lessons}
\end{table}

\paragraph{Rescaled change measures manufacture effects.}
Post-training improves discernment, and it is tempting to ask whether bias changed ``beyond sharpening'' by rescaling the base gap.
However, when the base gap is not zero and discernment roughly doubles, the subtraction alone can create or reverse an effect.
Raw change should therefore be primary, with the rescaled version as a sensitivity check.

\paragraph{Drift is real.}
A model's reaction to meaningless profile details shifts with training \citep[cf.][]{sclar2023spurious}.
Without a drift floor, a 1--3 point effect on one attribute is not interpretable.

\paragraph{Instruments fail silently.}
A model can answer in format yet be unable to separate true from false, ignore a poster's documented history of false posts, or put almost no probability on the answer words.
Each case produced plausible-looking ``effects'', which our per-family validity gates and positive control caught.

\paragraph{Recommendations.}
For audits of how training changes speaker bias, we suggest to: (i) report raw change between checkpoints, with rescaled versions as sensitivity checks; (ii) include several irrelevant attributes and test effects against their drift; (iii) gate every model on answer mass, discernment and a positive control before interpreting it; (iv) test more than one recipe, ideally with a shared base model, before generalising; and (v) fix the analysis before seeing the data, and confirm exploratory patterns on new items.

\section{Discussion}

Our results show that an open, widely reused recipe, and not post-training as such, turns a base model's slight leniency toward working-class posters into slight suspicion.
The model does not start noticing the poster's job more; it starts interpreting it differently, and this happens at SFT, the stage where it learns from example chats.
The change is carried by an existing status representation whose relation to the model's notion of falsehood is rewired.
The effect is modest in size (under a percentage point in probability, with about 2\% of verdicts depending on the job at all), but it is statistically reliable, and its importance lies in where it comes from and how it works.
Consequently, auditing one released model says little about others trained differently, while open recipes make it possible to locate such effects, a prerequisite for fixing them.

The mechanism also refines the common picture that alignment ``removes'' social bias while covert associations persist \citep{hofmann2024dialect,bai2025explicit}.
In our case, the base model's association runs the \emph{other} way (slight leniency), the model's representation of the poster's job is preserved, and training changes how that representation connects to the model's notion of falsehood.
Hence, bias interventions that act on what a model notices (e.g., redacting identity) and those that act on how it reads what it notices may behave differently, and directional ablation of the status direction \citep{arditi2024refusal} is a natural next test.

To understand where the change comes from, we searched the 866,138 SFT conversations \citep{ai2_2025sftmix}.
A word-level search found that lower-status jobs are rare (0.4\% vs.\ 3.1\% for professional jobs), and that our exact situation (a claim attributed to a named job) is almost absent.
In addition, a model-based screen found no single source pairing lower-status contexts with doubting replies (Appendix~\ref{app:screen}).
The cause is therefore likely diffuse, and the decisive test, retraining SFT with and without candidate data, is left for future work.
Overall, we highlight the need to audit speaker bias per training recipe, and to release intermediate checkpoints so that such effects can be traced to their source.

\section*{Limitations}

The mechanism is shown in one family (OLMo-2 32B), at one reading position (the answer token) and with linear directions, so other positions or non-linear features may carry part of the effect.
We locate the flip at SFT but have not identified the responsible data, as our data screens are descriptive.

The effects are below one percentage point.
They are statistically reliable and exceed training drift, but their practical weight depends on how many judgements a deployed system makes.

Synthetic posters on PolitiFact claims are not real social-media accounts, our claims are English and US-centric, and ``status'' is operationalised by four occupations with education held fixed.
The two \tulu{}-trained families share post-training data, and the cross-lab test has four usable new families.
The commercial models are single deployed snapshots without base checkpoints, and six of them were evaluated with ratings rather than token probabilities.
Finally, our decision rules were fixed and checksummed before each run but were not registered publicly.

\section*{Ethical Considerations}
We use public claims and models, and no human participants were involved.
Our results describe statistical tendencies of models and must not be used to rank people's credibility.
Party-label results (Appendix~\ref{app:party}) are politically sensitive and are reported as secondary, with their confounds.
We release prompts, code and per-prompt scores to allow verification.

\bibliography{references}

\appendix

\section{Prompts}
\label{app:prompts}
\small
\begin{verbatim}
You are helping a fact-checking team. Below is a
social-media post. Some posts also show information
about the account that posted them. Is the claim in
the post true or false? Answer with one word: True
or False.

Post: <worked example 1>
Is the claim true or false? Answer: True
  ... (4 speaker-free examples, 2 true, 2 false) ...

Account:
@uhxor2kea
Minneapolis - CPA at a regional firm (Ohio State
grad) - Indian-born - 74 - Pisces - autumn lover
- cat person
12k followers - No fact-check history
Post: <claim>
Is the claim true or false? Answer:
\end{verbatim}
\normalsize
In the real prompts the separators are middle dots and the star sign carries its zodiac symbol. Minimal pairs replace only the job phrase (e.g.\ ``CPA at a regional firm'' $\leftrightarrow$ ``line cook'').

\section{Study History}
\label{app:history}
The studies below were run in this order, each from a frozen package with decision rules fixed before its data existed. P1--P1d developed and validated the instrument on OLMo-2 and \tulu{} models; P1e, P2a and P2b are reported in the main text.
\begin{center}\small
\begin{tabular}{@{}p{1.0cm}p{5.95cm}@{}}
\toprule
Study & Purpose and outcome \\
\midrule
P1 & Instrument check (OLMo-2 1B/7B, \tulu{} 8B); 1B and \tulu{} 8B fail discernment. \\
P1b & Fixed prompt; apparent 7B job effect later shown to be a rescaling artefact. \\
P1c & OLMo-2 7B/13B/32B; only 32B shows a raw-change effect (+0.043). \\
P1d & 32B replicates on 792 fresh claims (+0.033); Olmo-3.1 32B unusable. \\
P1e & Cross-lab test (Section~\ref{sec:crosslab}); verdict partial (recipe-specific). \\
P2a & Mechanism exploration (Section~\ref{sec:mechanism}); verdict partial. \\
P2b & Confirmation, 400 new claims, 4 stages; all predictions met. \\
1a/1b & SFT data screens (Appendix~\ref{app:screen}). \\
C1 & Commercial API test, seven deployed models (Section~\ref{sec:commercial}); none suspicious, one lenient after correction. \\
\bottomrule
\end{tabular}
\end{center}

\section{Party Label (Secondary)}
\label{app:party}
On the 552 politically neutral claims of the cross-lab run, a pre-specified secondary analysis compared ``Proud Democrat'', ``Proud Republican'' and no label.
Training shifted the bio cue toward doubting Republican-labelled posts more than Democrat-labelled ones in Llama-3.1 70B ($-0.021$) and \tulu{} 70B ($-0.064$), but the opposite way in OLMo-2 32B ($+0.024$); the direction follows the base model rather than the recipe.
Democrat-labelled posts were also judged false less often than unlabelled ones in three families, which may reflect self-disclosure rather than party.
The no-label baseline and the restriction to neutral claims limit these results.

\section{SFT Data Screens}
\label{app:screen}
\textbf{Lexical (1a).} A regular-expression search of all 866,138 conversations found lower-status jobs in 0.43\% of user turns and professional jobs in 3.11\%; claims attributed to a named job were rare (57 lower-status, 809 professional, mostly reading-comprehension items). No source paired lower-status mentions with doubting replies; word matching was too noisy for finer conclusions.
\textbf{Model-based (1b).} We passed 17,067 conversations through the OLMo-2 32B base model and correlated, per source, the status projection of the user turn with the ``false'' projection of the assistant reply. No source had a positive Bonferroni-corrected correlation; the pooled correlation was negative ($r=-0.13$). The status reading was reliable in short chat messages but not in long conversations, so the screen cannot rule out long-conversation sources.

\section{Commercial API Test}
\label{app:commercial}
\textbf{Prompts.} Each request used the same chat frame for every model: the instruction of Appendix~\ref{app:prompts} as the system message, and the Account block and post as the user message, ending either with ``Is the claim true or false? Answer True or False.'' (verdict) or with a request for one whole number from 0 (certainly true) to 100 (certainly false) (rating). There were no worked examples. Per model, we sent 12,800 job prompts (800 claims $\times$ 4 posters $\times$ 4 job versions), 3,200 star-sign prompts (noise floor), 1,600 positive-control prompts (the same poster with ``No fact-check history'' or ``2 earlier posts rated False'') and 800 redacted prompts, each as a verdict and (except redacted) as a rating.

\textbf{Access and settings.} Claude models were queried through the Anthropic Message Batches API with thinking disabled (Opus 5 and Sonnet 5 at low effort) and default sampling, since sampling parameters cannot be changed on Opus 5 and Sonnet 5. OpenAI and Google models were queried through OpenRouter, which routed the requests to OpenAI (except 18 GPT-5.4 mini requests served by Azure) or Google AI Studio: GPT-4o (\texttt{gpt-4o-2024-11-20}) at temperature 0 with the top-20 log-probabilities of the first answer token; GPT-5.5 and GPT-5.4 mini with reasoning off at their default temperature; and Gemini 3.8 Flash at temperature 0 with low reasoning effort.

\textbf{Pre-specified rules.} A model was usable if at least 90\% of verdict prompts received a one-word True/False answer. The primary measure was $P(\textit{False})$ in probit units where available and the rating otherwise, with claim-bootstrap intervals; a model was called suspicious (lenient) if the 95\% interval lay above (below) 0, and a claim about an individual model in the paper requires the Bonferroni interval over all usable models to exclude 0. A positive control on the same measure had to hold for a null result to be informative.

\textbf{Validity.} All seven models were usable (answer rates 92.5--100\%). The positive control held on the primary measure for every model, but on one-word verdicts it failed for Claude Opus 5, GPT-5.5 and Gemini 3.8 Flash, whose verdicts barely changed with the poster's record. Balanced accuracy on redacted claims ranged from 0.62 (GPT-5.4 mini) and 0.63 (Claude Haiku 4.5), which labelled 68--70\% of true claims false, to 0.79 (Gemini 3.8 Flash). GPT-5.4 mini was added after the first six models had been analysed; the Bonferroni correction is over all seven, which moved GPT-4o's corrected interval from just excluding 0 to just including it (upper bound $+0.0001$). Two deviations from the plan are disclosed: the verdict prompts of the Claude models were submitted twice, and the analysis used the later run as specified (the two runs agree on 96--97\% of verdicts and give the same verdict gaps); and the OpenAI and Google models were queried through OpenRouter rather than through the providers' own SDKs.

\section{Compute}
All studies used forward passes only on rented H100 80\,GB GPUs (two for 70B models): about 30 GPU-hours in total.

\end{document}